# On the Relation between Kappa Calculus and Probabilistic Reasoning

Adnan Darwiche and Moisés Goldszmidt
Rockwell Science Center
444 High Street
Palo Alto, CA 94301
{darwiche,moises}@rpal.rockwell.com

## Abstract

We study the connection between kappa calculus and probabilistic reasoning in diagnosis applications. Specifically, we abstract a probabilistic belief network for diagnosing faults into a kappa network and compare the ordering of faults computed using both methods. We show that, at least for the example examined, the ordering of faults coincide as long as all the causal relations in the original probabilistic network are taken into account. We also provide a formal analysis of some network structures where the two methods will differ.

Both kappa rankings and infinitesimal probabilities have been used extensively to study default reasoning and belief revision. But little has been done on utilizing their connection as outlined above. This is partly because the relation between kappa and probability calculi assumes that probabilities are arbitrarily close to one (or zero). The experiments in this paper investigate this relation when this assumption is not satisfied. The reported results have important implications on the use of kappa rankings to enhance the knowledge engineering of uncertainty models.

## 1 Introduction

Bayesian reasoning has found widespread use in recent years [12]. Applications based on Bayesian networks, for example, have spanned over diagnosis, forecasting, natural language understanding, and planning [1]. But despite the popularity of Bayesian methods, one of their key aspects has always stood in their way to further success and wider use; namely, their commitment to point probabilities. In particular, most Bayesian techniques cannot commence without committing a domain expert to a full probability distribution, which typically requires many probabilities to be specified. Although recent advances in Bayesian networks have reduced this problem by appealing to conditional independences, there is still a significant interest in reducing this problem even further given its impact on knowledge elicitation and model building.

In recent years, a number of proposals have been extended for the purpose of relieving domain experts from having to specify point probabilities. Many of these proposals offer concrete methods that allow Bayesian reasoning to commence without a commitment to a complete probability distribution. An example of this is Qualitative Probabilistic Networks [18], which allow one to reason about probabilistic influences among variables in a qualitative manner that is consistent with Bayesian reasoning. A second class of proposals attempts to relief experts from providing point probabilities by requiring more abstract and intuitive belief measures that are consistent with point probabilities. A key proposal in this camp is kappa calculus [16, 17] and its probabilistic interpretation using $\epsilon$-semantics [8]. In this framework, experts can provide beliefs in the form of if-then rules that are quantified using order-of-magnitude probabilities. This quantification can be naturally embedded into a causal network, where the same set of Bayesian distributed algorithms can be applied [10, 2].

Both kappa calculus and its probabilistic interpretation have been extensively studied from the perspective of belief revision, nonmonotonic and defeasible reasoning [2, 8, 7, 10, 11, 16, 17, 15]. Kappa calculus was also proposed as a qualitative version of probabilistic reasoning in [9]. Yet, the formal relation between kappa calculus and probabilistic reasoning is established under the assumption that probabilities are extreme; that is, not only should they be close to one or zero but also they should be *arbitrarily* so. This requirement, which is never met in practice, means that kappa calculus can be viewed as an abstraction of probability calculus under the following *acceptance rule* [7]: Even though probabilities may not be arbitrarily extreme, the agent is willing to assume and behave as if they were, thus transforming them into plain beliefs quantified by kappa rankings that can be manipulated using kappa calculus.

The question we address in this paper is the follow-ing: What are the consequences of adopting the acceptance rule? For example, what information is lost once we are willing to take regular probabilities and abstract them into plain beliefs to be processed by kappa calculus? To answer these questions, we take an empirical approach and use a diagnostic example to test our hypothesis. Our results show that in spite of differences in absolute beliefs, when it comes to ordering the set of faults, both standard probabilities and their corresponding kappa rankings coincide most of the time. Moreover, an analysis of the differences between the two calculi led us to identify two causal structures where using probabilities or kappa rankings will yield different results.

The results in this paper are important for the knowledge engineering of uncertainty models for the following reasons:

1. Eliciting and building uncertainty models seems to be an easier task in kappa calculus than in probabilities. The kappa quantification of a network can be performed using if–then rules and ignorance can be specified by declaring that both an event and its negation are possible.

2. Models are more robust in kappa calculus, since small changes in the uncertainty will not affect much the assignment of beliefs.

3. It seems easier to absorb the results of a probabilistic inference once they are displayed as order-of-magnitude approximations (kappa rankings) of the actual probabilities.

4. There are indicators that algorithms based on extreme probabilities [13] and kappa rankings [9] can be faster than those based on regular probabilities.

This paper is structured as follows. We overview kappa calculus in Section 2 and then elaborate on its relation with probability calculus in Section 3. Specifically, although kappa calculus has been developed independently of probability calculus, kappa rankings can be viewed as order-of-magnitude probabilities when these probabilities are arbitrarily high or low. In Section 3, we provide a formal translation from point probabilities to kappa rankings and outline the role that this translation could play in practical systems, where probabilities are not necessarily extreme. We then report a number of experiments in Section 4 that are designed to evaluate the proposed translation and to assess the possible loss of information it could lead to. The experiments are conducted in the context of a diagnosis task. Some of the reported results lend themselves to formal analysis that we carry out in Section 5. The key outcome of this analysis is an intuitive characterization of kappa calculus on some of the causal structures appearing in real world applications. Finally, Section 6 summarizes the main results and offers another perspective on the connection between probabilities and kappa rankings according to which kappa rankings are strengths of default assumptions that are extracted from probabilistic information. This connection is in the spirit of earlier work on extreme probabilities and $\epsilon$-semantics [12] and provides a better understanding of the connection between point probabilities, kappa rankings and default priorities.

## 2   Kappa calculus

The original motivation behind kappa calculus was to propose a non-probabilistic theory of inductive reasoning [16, 17]. A non-probabilistic theory was sought because inductive reasoning involves classifying propositions according to whether they are believed or disbelieved and then changing this classification upon receiving further information. But classical probability theory did not support such a classification: propositions are only graded by their probabilities and are not classified into believed/disbelieved/uncommitted.

Given this motivation, the properties of kappa calculus can be justified without having to appeal to a probabilistic interpretation, which is how the calculus was argued for in [16]. There, a state of belief is represented by a ranking $\kappa$ that maps propositions into the class of ordinals such that

1. $\kappa(true) = 0$,
2. $\kappa(\alpha \vee \beta) = \min(\kappa(\alpha), \kappa(\beta))$.

A rule was also given for conditioning a state of belief $\kappa$ on evidence $\mu$:

$$\kappa(\alpha \mid \mu) = \kappa(\alpha \wedge \mu) - \kappa(\mu).$$

According to kappa calculus, a proposition $\alpha$ is believed to degree $s$ if $\kappa(\neg\alpha) = s$; is disbelieved to degree $s$ if $\kappa(\alpha) = s$; and is uncommitted if $\kappa(\alpha) = \kappa(\neg\alpha) = 0$. Moreover, the strengths of these beliefs decide which of them are retracted when accommodating a disbelieved evidence.

Kappa calculus then offers a framework for reasoning with defeasible beliefs, where the kappa rankings play the role of default priorities [10, 11]. But the calculus is analogous to probability calculus in the sense that it provides a similar machinery: a definition of a state of belief and a definition of conditionalization for accommodating evidence. This correspondence should not be surprising, however, given the symbolic generalization of probability theory in [4], which provides definitions for abstract states of belief and abstract conditionalization that subsume both probability and kappa calculi (see [14] also for a generalization of belief functions that subsumes kappa rankings).

## 3   Kappas and probabilities

Although Spohn has motivated kappa calculus as a theory of belief change, Spohn also noted the connection between kappas and nonstandard probabilities

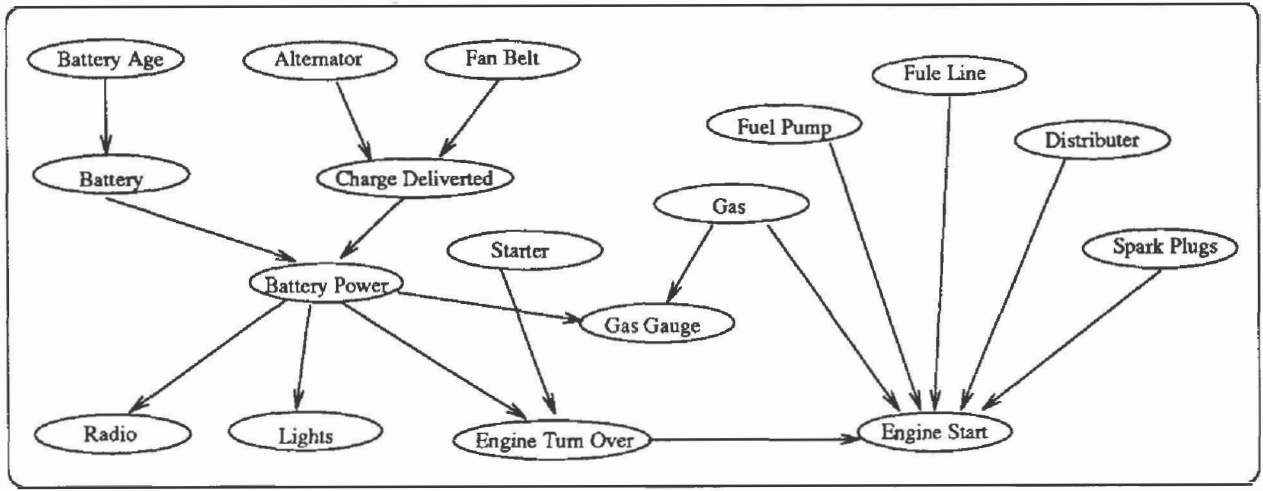

Figure 1: The car network.

[16, 17]. The purpose was mainly to explain the symmetry between properties of kappa calculus and laws of probability theory. In particular, Spohn suggested in [16] a mapping from probability distributions to kappa rankings that justifies the properties of kappa calculus. He proposed mapping each probability $Pr(\alpha \mid \beta)$ into a ranking $k$ such that $Pr(\alpha \mid \beta)/\epsilon^k$ is finite but not infinitesimal for an infinitesimal $\epsilon$. Spohn then showed that we get the following:

1. $\kappa(\alpha \vee \beta) = \min(\kappa(\alpha), \kappa(\beta))$,
2. $\kappa(\alpha \mid \beta) = \kappa(\alpha \wedge \beta) - \kappa(\beta)$,

which are the basic properties of kappa rankings. This result provides an interpretation of kappa rankings as order-of-magnitude approximations of probabilities through the following relation:

$$\epsilon < p/\epsilon^k \leq 1, \qquad (1)$$

which is equivalent to $\epsilon^{k+1} < p \leq \epsilon^k$.

This connection between kappa rankings and probabilities is of great theoretical interest. For example, its role has been explored at length in providing probabilistic semantics to defeasible if-then rules that are crucial to nonmonotonic reasoning [8]. But the connection between kappas and probabilities is also important from a purely probabilistic sense. That is, a key concern of Bayesian practitioners, for example, is to continue to enjoy the merits of Bayesian techniques while committing as little as possible to point probabilities. The view of kappa rankings as order-of-magnitudes probabilities is one way to satisfy this need. That is, instead of providing point probabilities, one provides kappa rankings. In fact, the role of such a connection goes beyond the knowledge elicitation process to at least two other areas:

1. Given probabilities that result from answering a query, we can map these probabilities into kappa rankings before we present them to experts or before we use them as inputs to other reasoning processes such as decision making.

2. Given a set of probabilities to be computed with, we can map these into kappas and then use kappa-specific algorithms for the computation. This step is significant if kappa-specific algorithms turn out to be more efficient than probabilistic ones, a hope that is being backed by recent results [13, 9].

One should emphasize though that the above connection between kappas and probabilities rests on assuming that $\epsilon$ is infinitesimal. Given a probability distribution, for example, the following two computations will yield the same results when an infinitesimal $\epsilon$ is used:

(C1) Computing posterior probabilities using probability calculus and then abstracting them into kappa rankings.

(C2) Abstracting probabilities into kappa rankings and then computing posterior kappa rankings using kappa calculus.

But unless probabilities are arbitrarily high or low, computations C1 and C2 will be equal in a trivial sense. For example, if all probabilities are known to be between .05 and .95, the mapping of Equation 1 will produce a zero kappa for each given probability. This means that the resulting kappa distribution will be trivial; all it says is that everything is possible and that nothing is believed or disbelieved.

Therefore, we are constrained in practice to select a noninfinitesimal $\epsilon$ to use in Equation 1. There will always be tension between how close the value of $\epsilon$ is to zero and how close the results of computations C1 and C2 will be. On one extreme, $\epsilon$ is very close to zero and the results of C1 equal those of C2 but possibly in a trivial sense because the generated kappa

1. If $p = 0$, then print $\infty$.
2. $k \leftarrow 0$.
3. $p \leftarrow p/\epsilon$.
4. If $p > 1$, then print $k$ otherwise $k \leftarrow k + 1$.
5. Goto 3.

Figure 2: A procedure for translating a probability value $p$ into a kappa value $k$ by finding a solution to the equation $\epsilon^{k+1} < p \leq \epsilon^k$.

rankings may have lost most of the probabilistic information. On the other extreme, $\epsilon$ is not close to zero, the resulting kappa rankings are not trivial, but the results could be different from those obtained using probability calculus. The experiments in the next section will assess the discrepancies between the results of kappa calculus and those of probability calculus when $\epsilon$ is not infinitesimal, using two different measures of discrepancy. Section 5 will then offer a formal analysis of these results by identifying cases in which such discrepancies are expected.

## 4 Experimental results

To empirically study the connection between kappa calculus and probabilistic reasoning in those instances were $\epsilon$ is not infinitesimal, we conducted a set of experiments with different values of $\epsilon$ and different evidence. These experiments were performed on a probabilistic causal network for diagnostic reasoning about faults in a car. The network is depicted in Figure 1.[1] Each experiment involved setting the value of $\epsilon$, providing observations in the form of evidence, evaluating the probabilistic network using probability calculus, translating the probabilistic network into a kappa network using the procedure in Figure 2, and then evaluating the resulting kappa network using kappa calculus.

We conducted three sets of experiments for $\epsilon = 0.2$, $\epsilon = 0.02$, and $\epsilon = 0.002$. We report below (see Tables 1 and 2) on the most representative results of the similarities and differences between kappa and probabilistic inference. The observables where **engine-start**, **gas-gauge**, **lights**, and **engine-turn-over**, while the faults where **alternator**, **battery**, **fuel-pump**, **gas**, **plugs** and **starter**. The value of **engine-start** was always set to **NO**.

To assess the discrepancies between kappa and probability computations, we used the following two criteria:

1. *Ordering of faults:* In Table 1 we order the faults according to their corresponding probabilities and kappas. The table contains eight "runs", where a run is defined by an instantiation of the evidence. The first line in each run corresponds to the ordering of faults according to their probabilistic de-grees of belief. The second line corresponds to the ordering of faults when $\epsilon = 0.2$ and the third line to $\epsilon = 0.02$.

[1] This network was obtained from the Bayesian group at Microsoft Research.

2. *Degrees of Belief:* In Table 2 we compare the probabilities of faults to their kappa rankings by transforming the posterior probabilities into kappa rankings following the procedure in Figure 2 and using $\epsilon = 0.02$.

The first criterion provides a practical measure of the correspondence between kappa and probabilistic inference when the kappa network is generated automatically from a probabilistic one. The second criterion is intended to compare the results of computations C1 and C2 in Section 3 when $\epsilon$ is not infinitesimal.

All of the experiments reported here were conducted using CNETS [3]: an experimental tool for representing and reasoning with generalized causal networks [2], which include kappa and probabilistic causal networks as special instances.

We have the following observations about the reported results:

### Ordering of the faults

When $\epsilon = 0.2$, the ordering of faults according to probabilities and kappas is the same in all the runs, provided we break ties in a particular manner. Ties in the kappa case are expected given that they represent an abstraction of the real probabilistic value.

When $\epsilon = 0.02$, the results are also very close, except that the most likely fault and the second most likely fault are inverted in runs 2,4 and 6. The discrepancies in these runs are due to the same reason: loss of information due to the kappa abstraction. In particular, the matrix quantifying **engine-starts** contains four rows in which the kappa of **engine-starts** and the kappa of its negation are both zeros. That is, there are four rows in which the matrix does not commit to whether **engine-starts** is believed or disbelieved. But when $\epsilon = 0.2$, the matrix of **engine-starts** commits to whether **engine-starts** is believed or disbelieved in each row.

### Degrees of belief in faults

Note that probabilities and kappas disagree more noticeably in belief strengths than in the ordering of faults. Kappas are generally much more committed to assign stronger beliefs to the possible existence of faults than probability. This property is illustrated in Figures 4, 5, and 6, where kappa beliefs are sharp and linear. These figures will be discussed in more detail in Section 5.

The disagreement of belief strength in Table 2 prompted the formal analysis in Section 5. We basically identified two causal structures, a chain and a fork — see Figure 3. The chain structure was motivated by the discrepancies on the de-grees of belief in the subnetwork involving the nodes alternator, charge-delivered and battery-power. The fork structure was motivated by the discrepancies involving the subnetwork composed of the nodes battery power, lights, radio, engine-turn-over, and gas-gauge.

| | GAS-GAUGE | LIGHTS | TURN-OVER | Calculi | Ordering | | | | | |
|---|---|---|---|---|---|---|---|---|---|---|
| 1 | NOT-EMPTY | WORK | YES | Pr | Fuel-Pump | Plugs | Alternator | Starter | Gas | Battery |
| | | | | $\kappa_{\epsilon=0.2}$ | Fuel-Pump$^?$ | Plugs | Alternator | Starter Gas Battery | | |
| | | | | $\kappa_{\epsilon=0.02}$ | Fuel-Pump$^+$ | Plugs Alternator | | Starter Gas Battery | | |
| 2 | EMPTY | WORK | YES | Pr | Gas | Fuel-Pump | Plugs | Alternator Starter | | Battery |
| | | | | $\kappa_{\epsilon=0.2}$ | Gas$^?$ | Fuel-Pump | Plugs | Alternator | Battery Starter | |
| | | | | $\kappa_{\epsilon=0.02}$ | Fuel-Pump$^+$ | Gas | Plugs Alternator | | Battery Starter | |
| 3 | NOT-EMPTY | WORK | NO | Pr | Fuel-Pump | Starter | Plugs | Alternator | Battery Gas | |
| | | | | $\kappa_{\epsilon=0.2}$ | Fuel-Pump | Starter Plugs | | Alternator | Battery Gas | |
| | | | | $\kappa_{\epsilon=0.02}$ | Fuel-Pump$^?$ | Starter Plugs | | Alternator | Battery Gas | |
| 4 | EMPTY | WORK | NO | Pr | Gas | Fuel-Pump | Starter | Plugs | Alternator | Battery |
| | | | | $\kappa_{\epsilon=0.2}$ | Gas$^?$ | Fuel-Pump | Starter Plugs | | Alternator | Battery |
| | | | | $\kappa_{\epsilon=0.02}$ | Fuel-Pump$^?$ | Gas Starter Plugs | | | Alternator | Battery |
| 5 | NOT-EMPTY | DONT | YES | Pr | Fuel-Pump | Alternator | Battery | Plugs | Gas Starter | |
| | | | | $\kappa_{\epsilon=0.2}$ | Fuel-Pump Alternator Battery | | | Plugs | Gas Starter | |
| | | | | $\kappa_{\epsilon=0.02}$ | Fuel-Pump$^+$ | Alternator$^?$ Battery$^?$ | | Plugs | Gas Starter | |
| 6 | EMPTY | DONT | YES | Pr | Gas | Fuel-Pump | Alternator | Battery | Plugs | Starter |
| | | | | $\kappa_{\epsilon=0.2}$ | Gas$^?$ | Fuel-Pump Alternator Battery | | | Plugs | Starter |
| | | | | $\kappa_{\epsilon=0.02}$ | Fuel-Pump$^+$ | Alternator$^?$ Battery$^?$ | | Gas | Plugs | Starter |
| 7 | NOT-EMPTY | DONT | NO | Pr | Alternator | Battery | Fuel-Pump | Plugs | Starter | Gas |
| | | | | $\kappa_{\epsilon=0.2}$ | Alternator$^?$ Battery$^?$ | | Fuel-Pump | Plugs | Starter | Gas |
| | | | | $\kappa_{\epsilon=0.02}$ | Alternator$^?$ Battery$^?$ Fuel-Pump$^?$ | | | Plugs | Starter | Gas |
| 8 | EMPTY | DONT | NO | Pr | Battery | Alternator | Fuel-Pump | Plugs | Gas | Starter |
| | | | | $\kappa_{\epsilon=0.2}$ | Battery$^?$ Alternator$^?$ | | Fuel-Pump | Plugs | Gas | Starter |
| | | | | $\kappa_{\epsilon=0.02}$ | Battery$^?$ Alternator$^?$ Fuel-Pump$^?$ | | | Plugs | Gas Starter | |

Table 1: Ordering of faults according to (1) posterior probabilities that resulted from evaluating the probabilistic car network and (2) posterior kappa rankings that resulted from evaluating the kappa car network. A "+" means believed and a "?" means uncommitted.

| GAS-GAUGE | LIGHTS | TURN-OVER | Battery | | Alternator | | Starter | | gas | | Fuel-Pump | | Plugs | |
|---|---|---|---|---|---|---|---|---|---|---|---|---|---|---|
| | | | $\kappa$ | Pr | $\kappa$ | Pr | $\kappa$ | Pr | $\kappa$ | Pr | $\kappa$ | Pr | $\kappa$ | Pr |
| NOT-EMPTY | WORK | YES | ok | ok | ok | ok* | ok | ok | ok | ok | bad | ? | ok | ok* |
| EMPTY | WORK | YES | ok | ok | ok | ok* | ok | ok | ok | ? | bad | ? | ok | ok* |
| NOT-EMPTY | WORK | NO | ok | ok | ok | ok* | ok | ok | ok | ok | ? | ? | ok | ok |
| EMPTY | WORK | NO | ok | ok | ok | ok* | ok | ok | ok | ? | ? | ? | ok | ok |
| NOT-EMPTY | DONT | YES | ? | ok | ? | ok | ok | ok | ok | ok | bad | ? | ok* | ok |
| EMPTY | DONT | YES | ? | ok | ? | ok* | ok | ok | ok | ? | bad | ? | ok* | ok |
| NOT-EMPTY | DONT | NO | ? | ? | ? | ? | ok | ok | ok | ok | ? | ? | ok | ok |
| EMPTY | DONT | NO | ? | ? | ? | ? | ok | ok | ok* | ok | ? | ? | ok | ok |

Table 2: Comparing (1) kappa rankings that are abstracted from posterior probabilities that resulted from evaluating the probabilistic car network to (2) posterior kappa rankings that resulted from evaluating the kappa car network. A "*" indicates a difference between the two kappa rankings. A "?" indicates that the kappa ranking of a fault and that of its negation were zeros, thus leading to ignorance about whether the fault is present.

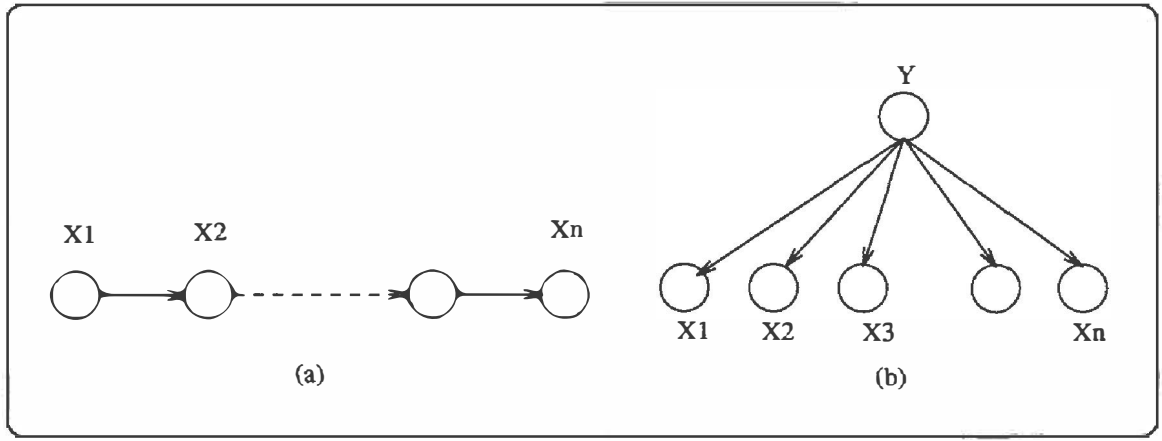

Figure 3: (a) Chain network, and (b) Fork network.

It is important to point out that even though the strength of belief between probabilities and kappas does not always coincide, the most plausible faults do agree. This suggests that the precise numbers may not be relevant if we dim them for the purposes of optimal recommendations regarding repair and actions.

## 5  Formal Analysis

The discrepancies we obtained in the previous experiments prompted the characterization of network configurations on which the use of kappa calculus leads to different results from probability calculus. In particular, we have identified two network structures where we can characterize such a discrepancy and analyze it intuitively. The first structure is that of a chain of variables and is discussed in Section 5.1. The second structure is that of a fork and is discussed in Section 5.2.

### 5.1  Propagation in chains

Consider the chain in Figure 3a, where all variables are assumed to be binary with values $x_i^+$ (*true*) and $x_i^-$ (*false*). Suppose that the causal links are quantified as follows: $Pr[x_1^+] = 0.5$, $Pr[x_i^+ \mid x_{i-1}^+] = 0.8$, and $P[x_i^+ \mid x_{i-1}^-] = 0.2$. Suppose further that we observe variable $X_1$ to be true. What can we conclude about the probability that a descendant $X_i$ is true?

According to probability calculus, the probability that any descendant $X_i$ is true will increase after observing that $X_1$ is true. Moreover, such increase will depend on how far $X_i$ is from $X_1$. In kappa calculus, however, we get a different behavior. That is, if we transform the previous probabilistic chain to a kappa chain using $\epsilon = 0.2$, we get the quantification: $\kappa[x_1^+] = \kappa[x_1^+] = 0$, $\kappa[x_i^- \mid x_{i-1}^+] = 1$ and $\kappa[x_i^+ \mid x_{i-1}^-] = 1$. Moreover, after observing that $X_1$ is true, each following $X_i$ will be believed true but with the same strength. That is, the strength of belief is independent of how far $X_i$ is from $X_1$, contrary to the probabilistic case. Figure 4 shows the difference between kappa and probability calculi with respect to the previous quantification of the chain.

In general though, the propagation of belief from variable $X_1$ to variable $X_j$ in such a network is governed probabilistically by the following equation:

$$P[x_i|x_1] = \sum_{x_2,\ldots,x_{i-1}} \prod_{k=2}^{i} P[x_k|x_{k-1}]. \qquad (2)$$

In contrast, kappa calculus leads to the following equation:

$$\kappa[x_i|x_1] = \min_{x_2,\ldots,x_{i-1}} \sum_{k=2}^{i} \kappa[x_k|x_{k-1}]. \qquad (3)$$

The kappa ranking corresponding to the result of Equation 2 should be equal to the result of Equation 3 when kappa rankings are generated using an infinitesimal $\epsilon$. In the experiments of Section 4, however, we used real valued $\epsilon$, which prompted the difference in the degrees of belief in Table 2.

### 5.2  Fusion in forks

Consider the network in Figure 3b, where all variables are also assumed to be binary. Suppose that the causal links are quantified as follows: $P[y^+] = 0.04$, $P[x_i^+|y^+] = 0.8$ and $P[x_i^+|y^-] = 0.2$. Suppose further that we observe variables $X_1$ through $X_i$ to be true. What can we conclude about the probability that $X_n$ is true?

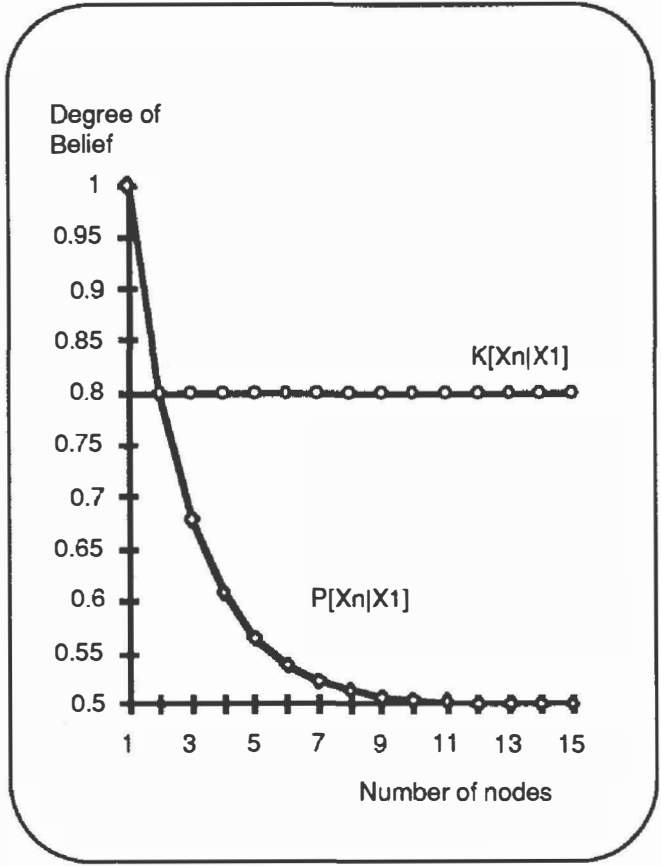

Figure 4: The horizontal axis represents $i-1$, the distance between variables $X_i$ and $X_1$ in Figure 3a. The vertical axis represents the belief that $X_i$ is true.

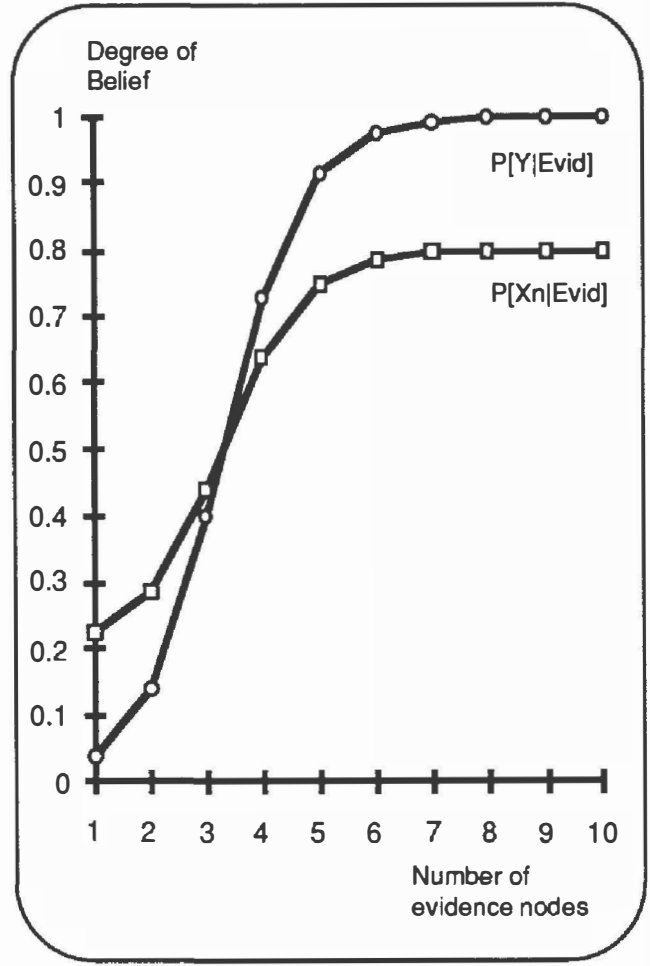

Figure 5: The horizontal axis represents $i$, the number of observed effects $X_1, \ldots, X_i$ in Figure 3b. The vertical axis represents the probabilities of $Y$ and $X_n$ being true.

In the probabilistic case, we expect that the previous evidence will increase the probability in $X_n$ being true. After all, the evidence increases the belief in $Y$ being true, which translates into an increase in the probability of $X_n$ being true. Moreover, the increase in the probability of $X_n$ depends on the number of observed variables $X_1, \ldots, X_i$. That is, the bigger $i$ is, the bigger the increase in the probability of $X_n$. Figure 5 supports this intuition by plotting the increase for a specific quantification of the network.

In the kappa case, however, observing the truth of effects $X_1, \ldots, X_i$ changes the belief in $X_n$ but in a different manner as depicted in Figure 6. That is, choosing the quantification: $\kappa[y^+] = 5$, $\kappa[x_i^-|y^+] = 1$ and $\kappa[x_i^+|y^-] = 1$ leads to the following. First, if the number of observations is less than five, $X_n$ is believed to be false. In case of six observations, $X_n$ is neither believed true nor false. But as we collect more observations, $X_n$ is then believed to be true, but the strength of this belief is not affected by the number of further observations.

The reason for this behavior stems from the following observation about kappa calculus. The belief in $x_n^+$ is affected by both the strength of $y^+$'s causal effect on $x_n^+$ and by the strength of believing in $y^+$. But the strength of believing in $y^+$ will be relevant only as long as it is no stronger than the causal effect. Once the belief in $y^+$ exceeds the strength of this causal effect, its exact value does not matter:

$$\kappa[x_n] = \min(\kappa[x_n \mid y^+] + \kappa[y^+], \kappa[x_n \mid y^-] + \kappa[y^-]),$$

which leads to $\kappa[x_n^-] = \min(1 + \kappa[y^+], \kappa[y^-])$ and $\kappa[x_n^+] = \min(\kappa[y^+], 1 + \kappa[y^-])$ in the above quantification. That is, if $y^+$ is unknown, then $x_n^+$ is unknown; if $y^+$ is believed to degree 1, $x_n^+$ is believed to degree 1; if $y^+$ is believed to degree 2, $x_n^+$ is believed to degree 1; and so on. Now, as we obtain more observations about the effects of $Y$, our belief in it increases, but that does not affect the belief in $X_n$ as shown above.

In general, the equations governing the propagation of belief both in the case of probabilities and kappas are

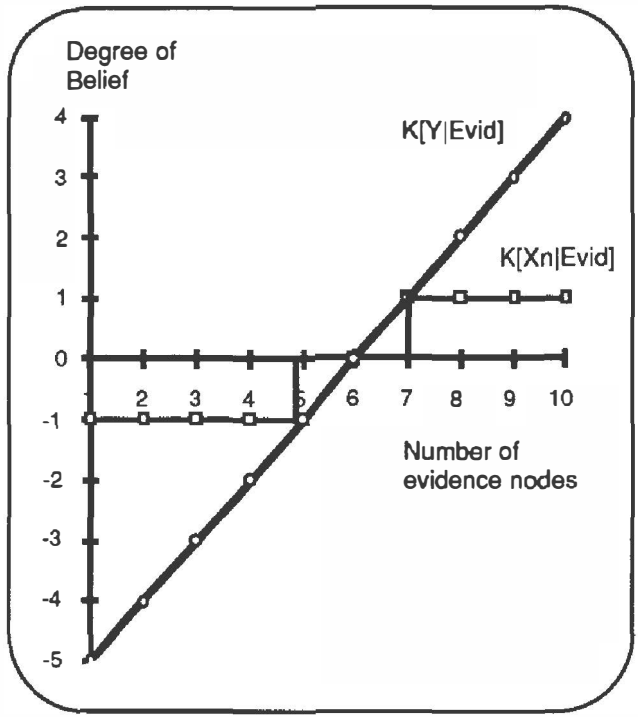

Figure 6: The horizontal axis represents $i$, the number of observed effects $X_1, \ldots, X_i$ in Figure 3b. The vertical axis represents $\kappa[z^-] - \kappa[z^+]$ for $z = y$ and $z = x_{10}$. If $\kappa[z^-] - \kappa[z^+]$ is positive, then $z^+$ is believed; if it is negative, then $z^+$ is disbelieved; otherwise, $z^+$ is unknown.

given below:

$$P[x_n|x_1, \ldots, x_i] = a \sum_y P[x_n|y] \prod_{k=1}^{i} P[x_k|y] P[y],$$

$$\kappa[x_n|x_1, \ldots, x_i] = b + \min_y \kappa[x_n|y] + \sum_{k=1}^{i} \kappa[x_k|y] + \kappa[y],$$

where $a$ and $b$ are normalization constants. As expected, in the case of probabilities, changes in the degree of belief of $X_n$ will be gradual and cumulative as new evidence on its sibling nodes is gathered. In contrast, the propagation of beliefs in the case of the kappa case will be abrupt and sharp.

## 6 Discussion

The use of probabilistic causal networks in diagnosis applications has become very common in recent years. One obstacle in this process, however, is the need to quantify causal relationships using point probabilities. Most often, probabilities are hard to assess and when they are provided, they seem to be too detailed for the reasoning tasks they are used to support. One possibility for simplifying this process is to quantify causal relationships using kappa rankings, thus inducing a kappa causal network that can be processed using kappa calculus. To adopt this practice, however, one must first provide answers to a number of questions. First, would kappa rankings keep us in the realm of probability theory, the properties of which have led to the popularity of probabilistic causal networks in the first place? Would kappa networks allow the same expressiveness that one expects from probabilistic causal networks? What should be done about the large body of existing probabilistic networks? Can these be mapped into kappa networks using some formal procedure? Would the resulting networks capture the information represented by the original probabilistic networks? And so on.

In this paper, we attempted to answer some of the above questions by (1) proposing a concrete mapping from probabilities to kappa rankings that does not require probabilities to be infinitesimal; (2) conducting an empirical study to assess the proposed mapping and to illustrate the expressiveness of kappa models in capturing diagnostic information; (3) providing some formal analysis of the connection between certain classes of probabilistic and kappa causal networks. The basic conclusion we have reached is that one may abstract a probabilistic network into a kappa network and still retain strong inferences. But our study also suggests that more needs to be said about when key inferences are retained.

The discrepancies obtained in inferences using probabilistic methods and kappa calculus should not be too surprising. Kappa calculus was proposed initially as a calculus for defeasible reasoning in which kappa rankings are interpreted as default priorities. As such, the calculus has been argued for convincingly in [16], has been shown to subsume many of the proposed calculi for defeasible reasoning in [6, 8], and has also contributed to the formalization of belief revision patterns that were not accounted for in the belief revision literature [5]. The calculus, therefore, seems to be very intuitive from a defeasible reasoning perspective and the inferences it leads to seem to be well justified. The discrepancies with probability calculus, and their relation, can then be explained as follows. Kappa calculus as a method for defeasible reasoning manipulates prioritized beliefs, which can be extracted from probabilistic information as suggested in Section 3. Yet, default priorities are less informative and capture less information than probabilities. Nevertheless, people seem to perform this kind of abstraction all the time, in spite of the possible loss of information. Most of our beliefs are probabilistic in nature but they get abstracted into default assumptions for various reasons, such as communicating them to others, indexing them efficiently, and simplifying their assessments. Thus, although the inferences made by kappa calculus can be well justified from the perspective of "plain beliefs" and "defeasible reasoning", they can disagree with probabilistic inferences.

The work in this paper takes the first steps towards answering questions of a bigger scope such as: When should we abstract probabilities into kappa rankings; for what purpose; and for what cost/gain? In this regard, we intend to continue this project in two directions. The first one concerns the process of decision making. The fact that the orders of faults were very similar in both probabilities and kappas suggests that the recommendations for repair should also be very similar. We intend to conduct a similar study to compare a probabilistic and a kappa decision making approach. The second direction concerns the computational value of abstracting probabilities into kappa rankings. The behavior of kappas in chains and forks suggest that the notion of belief acceptance in kappa calculus may yield a notion of *weak* independence where belief in a node may be enough to render other nodes independent in the network.[2] Our hope is that this property will translate into new algorithms with definite computational gains.

## Acknowledgments

We wish to thank J. Breese for his initial encouragement with this project, and M. Henrion and G. Provan for comments on a previous version of this paper.

---

[2] This possibility was also noted by one of the anonymous referees.